\pdfoutput=1

\documentclass[11pt]{article}

\usepackage{naacl2021}

\usepackage{times}
\usepackage{latexsym}

\usepackage[T1]{fontenc}

\usepackage[utf8]{inputenc}

\usepackage{microtype}

\usepackage{xurl}

\usepackage{multirow}
\usepackage{tabularx}
\usepackage{booktabs}
\usepackage{mathtools}
\usepackage{amssymb}
\usepackage{amsmath}
\usepackage{amsfonts}
\usepackage{bm}
\usepackage{bbm}
\usepackage{pifont}
\usepackage{caption}
\usepackage{subcaption}
\usepackage{adjustbox}

\usepackage{url}

\usepackage{tabto}
\usepackage{enumitem}
\usepackage{xspace}
\usepackage{lipsum} 

\usepackage{algorithm}
\usepackage{algorithmic}

\newcommand\copa[3]{
\begin{description}[noitemsep]
    \item \textbf{Premise}: #1
    \item \textbf{a)} #2
    \item \textbf{b)} #3
\end{description}
}

\newcommand\csexpl{Cs-Ex}

\newcommand{\learner}{f}

\newcommand{\easy}{\emph{easy}\xspace}
\newcommand{\hard}{\emph{hard}\xspace}

\title{Learning to Learn to be Right for the Right Reasons}

 \author{Pride Kavumba$^{1,2}$  \hspace{.9cm}
Benjamin Heinzerling$^{2, 1}$ \\ 
\bf{Ana Brassard$^{2,1}$} \hspace{.9cm}
 \bf{Kentaro Inui$^{1,2}$} \\
 $^1$Tohoku University \hspace{.2cm}
 $^2$RIKEN AIP \\
 \texttt{\{pkavumba, inui\} @ecei.tohoku.ac.jp } \\
 \texttt{\{benjamin.heinzerling,  ana.brassard\} @riken.jp } \\}

\begin{document}
\maketitle

\begin{abstract}
Improving model generalization on held-out data is one of the core objectives in commonsense reasoning.
Recent work has shown that models trained on the dataset with superficial cues tend to perform well on the \easy test set with superficial cues but perform poorly on the \hard test set without superficial cues.
Previous approaches have resorted to manual methods of encouraging models not to overfit to superficial cues.
While some of the methods have improved performance on \hard instances, they also lead to degraded performance on \easy instances.
Here, we propose to explicitly learn a model that does well on both the \easy test set with superficial cues and \hard test set without superficial cues.
Using a meta-learning objective, we learn such a model that improves performance on both the \easy test set and the \hard test set.
By evaluating our models on Choice of Plausible Alternatives (COPA) and Commonsense Explanation, we show that our proposed method leads to improved performance on both the \easy test set and the \hard test set upon which we observe up to 16.5 percentage points improvement over the baseline. 
\end{abstract}

\section{Introduction}
\label{sec:introduction}

Pre-trained language models such as BERT \cite{devlin-etal-2019-bert} and RoBERTa \cite{RoBERTa2019} have enabled performance improvements on benchmarks of language understanding \cite{wang2019superglue}.
However, improved performance is not only the result of increased ability to solve the benchmark tasks as intended, but also due to models' increased ability to ``cheat'' by relying on superficial cues \cite{gururangan-etal-2018-annotation, sugawara-etal-2018-makes, niven2019probing}.
That is, even though models may perform better in terms of benchmark scores, they often are \emph{right for the wrong reasons} \cite{mccoy2019right} and exhibit worse performance when prevented from exploiting superficial cues \cite{gururangan-etal-2018-annotation, sugawara-etal-2018-makes, niven2019probing}.

To analyze reliance on superficial cues and to evaluate methods that encourage models to be right for the \emph{right} reasons, i.e., to solve tasks as intended, training instances can be divided into two categories \cite{gururangan-etal-2018-annotation}:
\easy training instances contain easily identifiable superficial cues, such as a word that strongly correlates with a class label so that the presence or absence of this word alone allows better-than-random prediction \cite{niven2019probing}.
In contrast, \hard instances do not contain easily exploitable superficial cues and hence require non-trivial reasoning.
Models that exploit superficial cues are characterized by a performance gap: they show high scores on \easy instances, but much lower scores on \hard ones.

\begin{figure}[t]
 \centering
 \includegraphics[width=\linewidth]{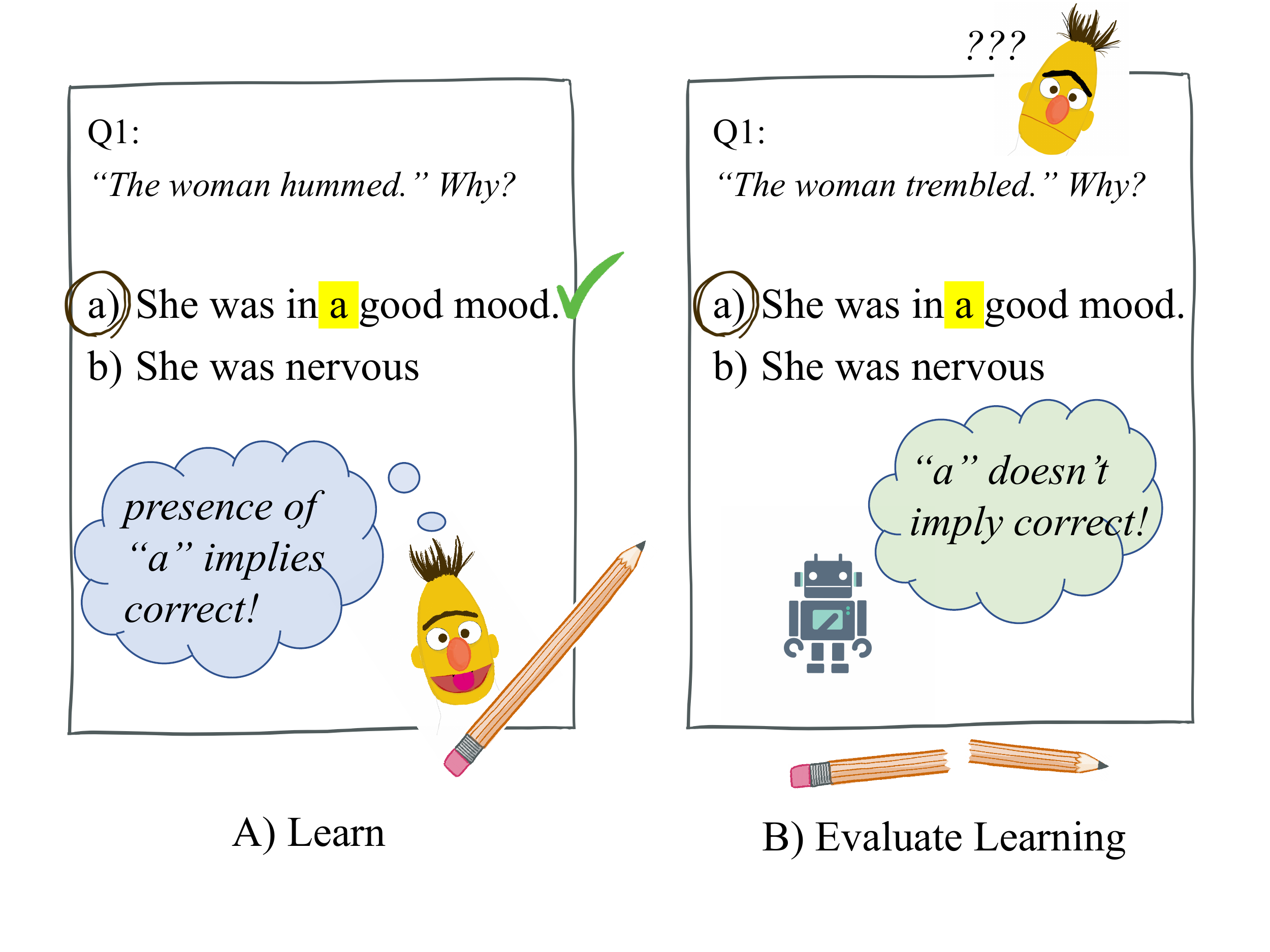}
 \caption{A modeling learning to be right for the right reason. A) shows a model wrongly learning that presence of ``a'' in the choice implies that the answer choice is correct. B) The models' learning is tested after a few examples and uses this testing error to improve how it learns in the inner-loop.
 Hence, learning to learn to be right for the right reasons.}
\end{figure}

Previous work has aimed at countering superficial cues.
A direct, if drastic, method is to completely remove \easy instances from the training data via adversarial filtering \cite{zellers-etal-2018-swag}, which leads to better performance on \hard instances, but, as \citet{gururangan-etal-2018-annotation} point out, filtering \easy instances may harm performance by reducing the data diversity and size.
Instead of completely removing \easy instances, \citet{schuster-etal-2019-towards} propose a loss discounting scheme that assigns less weight to instances which likely contain superficial cues, while \citet{belinkov-etal-2019-adversarial} use adversarial training to penalize models for relying on superficial cues.
A different approach, taken by \citet{niven2019probing} and \citet{kavumba-etal-2019-choosing}, is to augment datasets in a way that balances the distribution of superficial cues so that they become uninformative.
Common to all these approaches is that reduced reliance on superficial cues is reflected in degraded performance on \easy instances, while maintaining or increasing scores on \hard instances.

Here we propose meta-learning as an alternative approach to reducing reliance on superficial cues, which, as we will show, improves performance on both \easy and \hard instances.
Intuitively, we see reliance on superficial cues not as a defect of datasets, but as a failure to \emph{learn}: If a model learns to rely on superficial cues, it will not generalize to instances without such cues, but if the model learns \emph{not} to rely on such cues, this generalization will be possible.
Conversely, a model that only learns how to solve \hard instances may perform poorly on \easy instances.
Therefore, our meta-learned model learns how to generalize to both \easy and \hard instances.
By evaluating our method on two English commonsense benchmarks, namely Choice of Plausible Alternatives (COPA)~\cite{roemmele2011choice} and Commonsense Explanation (\csexpl{})~\cite{wang-etal-2019-make}, we show that meta-learning improves performance on both \easy and \hard instances and outperforms all baselines.

In summary, our contributions are:
\begin{enumerate}
    \item We propose a meta-learning method that learns how to generalize to both \easy and \hard instances (\S~\ref{sec:approach}),
    \item We show that Commonsense Explanation~\cite{wang-etal-2019-make} contain superficial cues that are easy to exploit by models (\S~\ref{sec:superficial_cues}),
    \item We empirically show that meta-learning a model to generalise to both \easy and \hard instances leads to better generalization not only on \hard instances but also on \easy instances (\S~\ref{sec:experiments}),
    
\end{enumerate}

\section{Learning to Generalize}
\label{sec:approach}

\subsection{Background}

Meta-learning has been successfully applied to problems such as few-shot learning~\cite{matching_NIPS2016_90e13578, maml-pmlr-v70-finn17a} and continual learning~\cite{JavedNEURIPS2019_f4dd765c, beaulieu2020learning}.

A meta-learning or learning to learn procedure consists of two phases. The first phase, also called meta-training, consists of learning in two nested loops. 
Learning starts in the inner loop where the models' parameters are updated using the meta-training training set.
At the end of the inner loop updates, the models' inner loop learning of the task is meta-train tested in outer loop where a separate meta-training testing set is used.
This is called meta-training testing.
Unlike a non-meta-training process, the meta-training testing error is also used to update the model parameters, i.e., the meta-training testing error is used to improve the inner loop.
Thus, learning is performed in both the inner and the outer loop, hence, learning-to-learn.

The second phase, also called meta-testing, consists only of a single loop. Model parameters are finetuned on a meta-testing training set and finally evaluated, only once, on the held out meta-testing testing set.
Note that the meta-testing testing set is different from the meta-training testing set.

One of the most popular meta-learning algorithms is Model-Agnostic Meta-Learning (MAML) algorithm~\cite{maml-pmlr-v70-finn17a}.
MAML is a few-shot optimization-based meta-learning algorithm whose goal is to learn initial model parameters $\theta$ for multiple related tasks such that a few gradient updates lead to optimal performance on target tasks.
We choose MAML for our experiments because it is model agnostic and, hence, widely applicable.

\subsection{Meta-Learning to Generalize}
\label{subsec:suml}

Our goal is to learn a model $f_{\theta}$, with parameters $\theta$, that generalizes well on both \easy instances, with superficial cues, and \hard instances, without superficial cues.
Specifically, given a large single-task training set $D^{tr}$, we want to be able to train a model that generalizes well to both the \easy test set $D^{test\_easy}$ and the \hard test set $D^{test\_hard}$.
To learn such a model, we require a meta-training testing set, $D^{tr_test}$, which contains both \easy and \hard instances.
Such a meta-training testing set will ensure that we evaluate the model generalization to both \easy and \hard instances.
Optimizing only for better performance on \hard instances can lead to poor generalization to \easy instances~\cite{gururangan-etal-2018-annotation}.

We cannot naively apply the meta-learning method designed for learning multiple few-shot tasks to a large dataset. 
A large dataset presents three main challenges.
First, a naive meta-learning method would require using the entire training set during each inner loop update.
This would make training very slow and computationally expensive.
To address this problem, we use small randomly sampled batches in each inner loop.
This is similar to treating each mini-batch as a single MAML task.

Second, a naive meta-learning method would require using the entire meta-training testing set for each outer loop update.
This, too, would make learning very slow when the meta-training testing set is large.
We address this challenge by evaluating the inner loop learning using only a small batch that is randomly drawn from the meta-training testing set.

Third, a naive meta-learning method would require storing the entire inner loop computation graph to facilitate second-order gradients' computation.
However, for large datasets and large models, such as recent pre-trained language models used in this paper, this is computationally too expensive and impractical on current hardware.
To address this problem, we use first-order MAML~\cite{maml-pmlr-v70-finn17a} that uses only the last inner-update.

We call this method of using random meta-training training batches and meta-training testing batches for meta-updates as Stochastic-Update Meta-Learning (SUML, Algorithm 1).
The hyperparameter k is the number of inner loop updates performed for each outer loop update (i.e., i in Algorithm 1 ranges from 1 to k).
Setting the value of k to 1 would make training unstable, much like using a batch size of 1 in standard (non-meta) training. On the other hand, a large value of k would make training slow.

Effectively, the model is meta-trained to use any batch in the training set to perform well on both the \easy and the \hard instances.

\begin{algorithm}[h]
\caption{Stochastic-Update Meta-Learning}
\label{alg:maml}
\begin{algorithmic}[1]
\REQUIRE $D^{tr}$: Training set
\REQUIRE $D^{tr\_test}$: Balanced meta-training test set
\REQUIRE $\alpha$: inner-loop step size
\REQUIRE $\beta$: outer-loop step size
\STATE $\theta$: LM pretrained parameters 
\WHILE{not done}
\STATE Sample batch $D_j^{tr\_test}$ from $D^{tr\_test}$
  \FOR {$i=1,2,3,...,k$}
  \STATE Sample batch $D_i^{tr}$ from $D^{tr}$
 \STATE Adapt parameters with gradient descent: $\theta_{i}=\theta_{i-1}-\alpha \nabla_\theta  L(  \learner_{\theta_{i-1}}, D_i^{tr} )$
 \ENDFOR
 \STATE Update $\theta \leftarrow \theta - \beta \nabla_\theta L( \learner_{\theta_k}, D_j^{tr\_test})$
\ENDWHILE
\end{algorithmic}
\end{algorithm}

\section{Superficial Cues in COPA and Commonsense Explanation}
\label{sec:superficial_cues}

\subsection{Datasets}
\label{subsec:datasets}

Here, we briefly describe the English commonsense datasets that we use in this paper.

\begin{table*}[t]
\centering
\adjustbox{max width=\textwidth}{%
\begin{tabular}{@{}lllccc@{}}
\toprule
Model    &  Ensemble & Data         &  \multicolumn{1}{l}{Easy} & \multicolumn{1}{l}{Hard} & \multicolumn{1}{l}{Overall} \\ \midrule
SuperGlue Leaderboard Score: \\
\quad RoBERTa-large + ALBERT-xxLarge & yes & COPA & - & - & 90.8 \\
\quad RoBERTa-large & yes &  COPA & - & - & 90.6 \\  \cmidrule(lr){1-6}
RoBERTa-large & no & COPA  & 90.5 & 83.9 & 86.4 \\
RoBERTa-large-adversarial  & no & COPA & 70.0 & 56.5 & 61.6  \\
RoBERTa-large + Balanced data & no & B-COPA  & 90.0   & 88.1 & 88.8 \\ 
RoBERTa-large-meta-learned \emph{(ours)} & no & COPA  & \textbf{92.6}       & \textbf{89.7} & \textbf{90.8} \\ \midrule
RoBERTa-base  & no & COPA & 80.0 & 71.3 & 74.6  \\
RoBERTa-base-adversarial & no & COPA & 76.3 & 57.0 & 64.4   \\
RoBERTa-base & no &  B-COPA           &  78.4          &  78.1          &  78.2  \\ 
RoBERTa-base-meta-learned \emph{(ours)} & no &  COPA         &  \textbf{87.9} &  \textbf{78.4} &  \textbf{82.0}               \\\cmidrule{1-6}\morecmidrules\cmidrule{1-6}
RoBERTa-large & no & Cs-Ex      & 98.0          &  79.1          &  93.8  \\
RoBERTa-large-adversarial & no & Cs-Ex & 94.0 & 59.0 & 86.2 \\ 
RoBERTa-large-meta-learned \emph{(ours)} & no &  Cs-Ex  &   \textbf{98.9} &  \textbf{87.1} &  \textbf{96.2}               \\ \midrule
RoBERTa-base  & no & Cs-Ex      & 95.0          &  62.1          &  87.7  \\
RoBERTa-base-adversarial & no & Cs-Ex & 93.8 & 54.8 & 85.2 \\
RoBERTa-base-meta-learned \emph{(ours)} & no & Cs-Ex   &   \textbf{97.6} &  \textbf{78.6} &  \textbf{93.4}               \\ \bottomrule
\end{tabular}
} 
\caption{Accuracy on Easy and Hard instances. We report accuracy for models trained on COPA, Balanced COPA (B-COPA) and Commonsense Explanation (Cs-Ex). We also report SuperGlue~\cite{wang2019superglue} leaderboard scores for single task fine-tuning for reference. }
\label{tab:main-results}
\end{table*}

\textbf{Balanced COPA}: The Balanced Choice of Plausible Alternatives~\cite[Balanced COPA]{kavumba-etal-2019-choosing} counters superficial cues in the answer choices of the Choice of Plausible Alternatives~\cite[COPA]{roemmele2011choice} by balancing token distribution between correct and wrong answer choices. Balanced COPA creates mirrored instances for each of the original instance in the training set. Concretely, for each original COPA instance shown below:

\copa
{The stain came out of the shirt. What was the CAUSE of this?}
{I bleached the shirt. (Correct)}
{I patched the shirt.}

Balanced COPA creates another instance that shares the same alternatives but a different manually authored premise.
The wrong answer choice the original question is made correct by the new premise (refer to App.~\ref{app:datasets} for more examples).
\copa
{\textit{The shirt did not have a hole anymore}. What was the CAUSE of this?}
{I bleached the shirt.}
{I patched the shirt. (Correct)}
This counters superficial cues by balancing the token distribution in the answer choices.

\noindent
\textbf{Commonsense Explanation}: Commonsense Explanation (\csexpl{})~\cite{wang-etal-2019-make} is a multiple-choice benchmark that consists of three subtasks. Here we focus on a commonsense explanation task.
Given a false statement such as \emph{He drinks apple.}, \csexpl{} requires a model to pick the reason why a false statement does not make sense, in this case either: a) \emph{Apple juice are very tasty and milk too}; or b) \emph{Apple can not be drunk} (correct); or c) \emph{Apple cannot eat a human}.

\subsection{Superficial cues in \csexpl{}}
While COPA has already been shown to contain superficial cues by \citet{kavumba-etal-2019-choosing}, \csexpl{} has not been analyzed yet. Here, we present an analysis of superficial cues in \csexpl{}.

We fine-tuned RoBERTa-base and RoBERTa-large with contextless inputs (answers 
only).
This reveals the models' ability to rely on shortcuts such as different token distributions in correct and wrong answers
\cite{gururangan-etal-2018-annotation, mccoy2019right}.

In this setting, we expect the models' accuracy to be nearly random if the answer choices have no superficial cues.
But, we find that RoBERTa performs better than random accuracy of 33.3\%.
The above-random performance of RoBERTa-base (82.1\%) and RoBERTa-large (85.4\%) indicates that the answers of \csexpl{} contain superficial cues.

\label{sec:unigram-cues}
To identify the actual superficial cues a model can exploit, we collect words/unigrams that are predictive of the correct answer choice using the \emph{productivity} measure introduced by \citet[][see definition in App.~\ref{app:productivity}]{niven2019probing}.
Intuitively, the productivity of a token expresses how precise a model would be if it based its prediction only on the presence of this token in a candidate answer.
We found that the word \emph{not} was highly predictive of the correct answer, followed by the word \emph{to} (See details in App.~\ref{app:productivity}).

\subsection{Easy and Hard Instances}
Following previous work~\cite{gururangan-etal-2018-annotation,kavumba-etal-2019-choosing}, we split the test set of \csexpl{} into an \easy and \hard subset. 
The \easy subset consists of all instances (1,572) that RoBERTa-base solved correctly across three different runs in the contextless input (answer only) setting. 
All the remaining instances, 449, are considered \hard instances.
For COPA, we use the \easy and \hard subset splits from \citet{kavumba-etal-2019-choosing}, which consists of 190 \easy and 310 \hard instances.

\section{Experiments}
\label{sec:experiments}

\subsection{Training Details}

In our experiments, we used a recent state-of-the-art large pre-trained language model, namely RoBERTa~\cite{RoBERTa2019}---an optimized variant of BERT~\cite{devlin-etal-2019-bert}.
Specifically, we used RoBERTa-base and RoBERTa-large with 110M and 355M parameters, respectively, from the publicly available Huggingface source code~\cite{Wolf2019HuggingFacesTS}.~\footnote{\url{https://github.com/huggingface/transformers}}
We ran all our experiments on a single NVIDIA Tesla V100 GPU with 16GB memory.

We used an Adam optimizer~\cite{kingma2014adam} with a warm-up proportion of 0.06 and a weight decay of 0.01.
We randomly split the training data into training data and validation data with a ratio of 9:1.
We trained the models for a  maximum of 10 epochs with early stopping based on the validation loss (full training details in App.~\ref{app:training-details}).

\subsection{COPA}
\label{subsec:balance-copa-experiments}

To evaluate the effectiveness of meta-learning a model to be robust against superficial cues,
we compare our model that is meta-trained on 450 original COPA instances and 100 balanced meta-training testing examples with three different baselines. 
Specifically, we compare to: \newline
1. A model trained on 500 original COPA instances. \newline
2. An adversarial trained model to avoid the answer only superficial cues on 500 original COPA instances. \newline
3. A model trained on 1000 Balanced COPA instances, manually created to counter superficial cues. 
In comparison, our meta-trained model uses only a small fraction of balanced instances. Effectively, our method replaces the need to have a large balanced training set with a small, 100 instances, in this case, meta-training test set.

The results show that the models trained on the original COPA perform considerably better on the \easy subset (90.5\%) than on the \hard subset (83.9\%)~(Table~\ref{tab:main-results}).
The models trained on balanced COPA improves performance on the \hard subset (88.1\%) but slightly degrades performance on the \easy subset (90.0\%).
This indicates that training on Balanced COPA improves generalization on the \hard instances.
As expected, the performance of the adversarial trained model is lower than the vanilla baselines.
This finding is similar to the result found in natural language inference~\cite{belinkov-etal-2019-adversarial}.
Comparing our meta-trained models to the baselines, we see that meta-training improves performance on both the \easy subset and \hard subset.
Our meta-trained models even outperform the models trained on nearly twice the training data and an ensemble of RoBERT-large.
It even matches an ensemble of RoBERTa-large and ALBERT-xxlarge~\cite{albert2019lan}.~\footnote{The SuperGlue leaderboard, from which the results shown in the first two rows of Tab.~\ref{tab:main-results} were taken, does not publish system outputs, so it’s not possible to compute scores on \easy and \hard subsets. And, the ensemble models reported have not been published yet, and there is no paper or source code which describes the model and training procedure, so it is not possible to reproduce these results.}

\subsection{Commonsense Explanation}
\label{subsec:commonsense-explanation-experiments}

This experiment aims to investigate an automatic method of creating a meta-training testing set.
Here we assume that there is no budget for manually creating a small meta-training testing set as in Balanced COPA.
We created a meta-training testing set by randomly sampling 288 \hard instances.
\citet{gururangan-etal-2018-annotation} pointed out that optimizing only for \hard instance might lead to poor performance on \easy instance.
This observation motivates us to include both \easy and \hard instances in the meta-training testing set, with the expectation that this will ensure that performance on \easy instances does not degrade.
We augmented the \hard instances with an equal number of randomly sampled \easy instances, resulting into the final meta-training testing set with 576 instances.

The results show that the meta-trained models perform better than the baselines on both \easy and \hard instances (Table~\ref{tab:main-results}).
For RoBERTa-large we see 0.9 percentage point improvement on \easy instances and eight percentage points improvement on the \hard instances.
We see the largest gains on the RoBERTa-base with 2.6 and 16.5 percentage points on \easy and \hard instances, respectively.
The results indicate that in the absence of a manually authored meta-training testing set without superficial cues, we can use a combination of \easy and \hard instances.

\section{Conclusion}
We propose to directly learn a model that performs well on both instances with superficial cues and instances without superficial cues via a meta-learning objective.
We carefully evaluate our models, which are meta-learned to improve generalization, on two important commonsense benchmarks, finding that our proposed method considerably improves performance across all test sets.

%

\section*{Acknowledgements}
This work was supported by JST CREST JPMJCR20D2. 

\bibliography{custom}
\bibliographystyle{acl_natbib}

\clearpage
\newpage
\appendix
\section*{Appendix}

\section{Identifying Superficial Cues}
\label{app:productivity}
We identify tokens predictive of the correct answer using \emph{productivity}, as defined by \citet{niven2019probing}. Let $\mathbb{T}_{j}^{(i)}$ be the set of tokens in the alternatives for data point ${i}$ with label ${j}$.
The \emph{applicability} ${\alpha}_{k}$ of a token $k$ counts how often this token occurs in an alternative with one label, but not the other:
$$\alpha_{k}=\sum_{i=1}^{n} \mathbbm{1}\left[\exists j, k \in \mathbb{T}_{j}^{(i)} \wedge k \notin \mathbb{T}_{\neg j}^{(i)}\right]$$
The \emph{productivity} $\pi_k$ of a token is the proportion of applicable instances for which it predicts the correct answer:
$$\pi_{k}=\frac{\sum_{i=1}^{n} \mathbbm{1}\left[\exists j, k \in \mathbb{T}_{j}^{(i)} \wedge k \notin \mathbb{T}_{\neg j}^{(i)} \wedge y_{i}=j\right]}{\alpha_{k}}$$

The most productive tokens in \csexpl{}shown in Table~\ref{tab:statistical-cues-table}.

\section{Dataset}
\label{app:datasets}
We use English datasets, namely COPA~\footnote{\url{https://people.ict.usc.edu/~gordon/downloads/COPA-resources.tgz}}~\cite{roemmele2011choice}, Balanced COPA~\footnote{\url{https://balanced-copa.github.io/}}~\cite{kavumba-etal-2019-choosing},  and Commonsense Explanation~\footnote{\url{https://github.com/wangcunxiang/SemEval2020-Task4-Commonsense-Validation-and-Explanation}}~\cite{wang-etal-2019-make} for all our experiments.

\subsection{Balanced COPA}
The Balanced Choice of Plausible Alternatives~\cite[Balanced COPA]{kavumba-etal-2019-choosing} counters superficial cues in the answer choices by extending the training set of Choice of Plausible Alternatives~\cite[COPA]{roemmele2011choice} with twin questions for each of the original COPA instances.
Examples of twin questions are shown below:

\noindent
\textbf{Example 1:} \newline
Original instance
\copa
{My body cast a shadow over the grass. What was the CAUSE of this?}
{The sun was rising. (Correct)}
{The grass was cut.}
Mirrored instance:
\copa
{\textit{The garden looked well-groomed.}. What was the CAUSE of this?}
{The sun was rising.}
{The grass was cut. (Correct)}

\noindent
\textbf{Example 2:} \newline
Original instance
\copa
{The woman tolerated her friend's difficult behavior. What was the CAUSE of this?}
{The woman knew her friend was going through a hard time. (Correct)}
{The woman felt that her friend took advantage of her kindness.}
Mirrored instance:
\copa
{\textit{The woman did not tolerate her friend's difficult behavior anymore.}. What was the CAUSE of this?}
{The woman knew her friend was going through a hard time.}
{The woman felt that her friend took advantage of her kindness. (Correct)}

\subsection{Commonsense Explanation}
Commonsense Explanation (\csexpl{})~\cite{wang-etal-2019-make} is a multiple-choice benchmark that consists of three subtasks. Here we focus on a commonsense explanation task.
Given a false statement, \csexpl{} requires a model to pick the reason why a false statement does not make sense.
For example:
\begin{description}[noitemsep]
    \item \textbf{FalseStatement:} He drinks apple.
    \item \textbf{a)} Apple juice are very tasty and milk too
    \item \textbf{b)} Apple can not be drunk (correct)
    \item \textbf{c)} Apple cannot eat a human
\end{description}

\section{Training Details}
\label{app:training-details}

In our experiments, we use a state-of-the-art recent pre-trained language model, namely RoBERTa~\cite{RoBERTa2019}, an optimized variant of BERT~\cite{devlin-etal-2019-bert}.
We use RoBERTa-base and RoBERTa-large with 110M and 355M parameters respectively from the publicly available Huggingface source code~\cite{Wolf2019HuggingFacesTS}.~\footnote{\url{https://github.com/huggingface/transformers}}.
We run all our experiments on a single NVIDIA Tesla V100 GPU with 16GB memory.

We use Adam~\cite{kingma2014adam} with a warm-up proportion of 0.06 and a weight decay of 0.01.
We randomly split the training data into training data and validation data with a ratio of 9:1.
We train the models for a  maximum of 10 epochs with early stopping based on the validation loss.

\subsection{COPA Baselines}

\begin{table}[t]
\centering
\adjustbox{max width=\linewidth}{%
\begin{tabular}{@{}llcccccc@{}}
\toprule
\multirow{2}{*}{Dataset}        & \multirow{2}{*}{Word} & \multicolumn{2}{c}{Train} & \multicolumn{2}{c}{Dev}  \\ \cmidrule(l){3-4} \cmidrule(l){5-6} 
                                &                       & Prod.        & Cov.       & Prod.       & Cov.              \\ \midrule
\multirow{2}{*}{\csexpl} & not                   & \textbf{72}         & 47       & \textbf{59}        & 40              \\  
                                & to                    & 35         & 37       & 33        & 43              \\  \bottomrule
\end{tabular}
}
\caption{Productivity (Prod.) and Coverage (Cov.) of the top 2 most productive tokens in each dataset. We have highlighted the most productive results. In Commonsense-Explanation, if one always picks an option with `not' then one can achieve 59\% with a coverage of 40\% data points on the dev set.}
\label{tab:statistical-cues-table}
\end{table}

We use grid search for hyperparameter from learning rates \{1e-5, 8e-6, 6e-6, 4e-6, 2e-6, 1e-6\}, batch sizes \{4, 8, 16, 32, 64\}, gradient accumulation \{1, 2, 4, 8\}, Adam $\beta_2$ \{0.98, 0.99\}, and with gradient norm clipping of 1 and no gradient norm clipping.
We pick the best performing hyperparameters on the validation set.

\subsection{Commonsense Explanation (Cs-Ex) Baselines}

We test learning rates 1e-5, 8e-6, 6e-6, 4e-6, 2e-6 and 1e-6, Adam $\beta_2$  0.99, and with gradient norm clipping of 1.
For RoBERTa-base, we use batch sizes of 64 with gradient accumulation 1, and for RoBERTa-large, we use a batch size of 32 with gradient accumulation 2. 
We pick the best performing hyperparameters on the validation set.

\subsection{Adversarial Trained Baseline}
We follow the setup defined by \citet{belinkov-etal-2019-adversarial}.
Specifically we optimize the objective function:

$\begin{aligned} 
L &=L_{\mathrm{scorer}}+\lambda_{\mathrm{Loss}} L_{\mathrm{Adv}} \\ L_{\mathrm{Adv}} &=L\left(c_{\mathrm{choice}}\left(\lambda_{\mathrm{Enc}} \operatorname{GRL}_{\lambda}\left(g_{C}(C)\right), y\right)\right) 
\end{aligned}$
Where $L_{\mathrm{scorer}}$ is the loss of the multiple-choice scorer (or head), $GRL_{\lambda}$ is the gradient reversal layer~\cite{pmlr-v37-ganin15}, $\lambda_{\mathrm{Loss}}$ is the importance of the adversarial loss ($L_{\mathrm{Adv}}$), $\lambda_{\mathrm{Enc}}$ is the scaling factor that multiplies the gradients after reversing them, $c_{\mathrm{choice}}$ maps the answer choice representation $C$ to an output y.
The goal is to obtain a representation $g_{C}(C)$ so that it is maximally informative for multiple-choice answering while simultaneously minimizes the ability of $c_{\mathrm{choice}}$ to accurately predict the correct choice (refer to \citet{belinkov-etal-2019-adversarial} for further details). 
We use grid search to tune hyperparameters $\lambda_{\mathrm{Enc}}$ \{0.1, 0.2, 0.3, 0.4, 0.5, 0.6, 0.7, 0.8, 0.9, 1 \} and $\lambda_{\mathrm{Loss}}$ \{0.1, 0.2, 0.3, 0.4, 0.5, 0.6, 0.7, 0.8, 0.9, 1 \} and report the best performing results on the development set.

\subsection{Meta-Training}

In the inner-loop, we pick the maximum batch size that fits in GPU memory.
For all the experiments, we use vanilla Stochastic Gradient Descent for the inner-loop with learning rate $\alpha$ 0.01 (it worked well in the first run therefore we do not modify it for the rest of the experiments), and Adam for the outer-loop with learning rate $\beta$ 1e-5 (based on the best learning rate for the RoBERTa baseline).

\end{document}